\documentclass[letterpaper, 10 pt, conference]{ieeeconf}  

\IEEEoverridecommandlockouts                              

\usepackage{graphics} 
\usepackage{graphicx}
\usepackage{xcolor}
\usepackage{amsmath} 
\usepackage{amssymb}  

\title{\LARGE \bf
WayFinder: Hierarchical Visual-Language-Action for Zero-Shot Waypoint Generation and Low-Level Kinematic Control
}

\author{
    Timothy K Johnsen$^{1}$ and Marco Levorato$^{2}$
\thanks{
    $^{1}$Timothy Johnsen is a PhD candidate in the Computational Science joint program with San Diego State University and The University of California, Irvine in CA, USA.
        {\tt\small tjohnsen@uci.edu}
}
\thanks{
    $^{2}$Marco Levorato is a full professor in the Computer Science department at the University of California, Irvine in CA, USA.
        {\tt\small levorato@uci.edu}
}
}

\usepackage{fancyhdr}
\fancypagestyle{arxiv}{%
    \fancyhf{}
    \chead{\footnotesize Accepted to the 2026 IEEE/RSJ International Conference on Intelligent Robots and Systems (IROS 2026).}%
}

\begin{document}

\bstctlcite{IEEEexample:BSTcontrol}

\maketitle
\thispagestyle{arxiv}
\pagestyle{empty}

\begin{abstract}

Visual–Language–Action (VLA) models offer unprecedented generalization for autonomous robots; however, their real-world deployment is frequently bottlenecked by unreliable execution and the prohibitive computational cost of fine-tuning for specific robot embodiments and tasks. To bridge this gap, we propose WayFinder, an end-to-end, closed-loop hierarchical VLA framework that circumvents the need for fine-tuning by decoupling high-level task reasoning from low-level kinematic control. WayFinder utilizes a zero-shot, offboard Multimodal Large Language Model (MLLM) policy to process linguistic context and state maps for strategic waypoint generation. Asynchronously, a lightweight, onboard policy executes real-time kinematic control at high frequency based on continuous sensor feedback. We evaluate WayFinder in Microsoft AirSim, testing on four environments of varying complexity and three MLLM scales to balance prediction efficacy with computational efficiency. Our results demonstrate that WayFinder achieves superior navigation reliability compared to baseline low-level policies. By querying the high-level MLLM only during navigation failures, WayFinder eliminates the need for fine-tuning, minimizes expensive inferences, and significantly increases navigation success rates by up to 27.45\%.

\end{abstract}

\section{INTRODUCTION}

The push toward fully autonomous robotics, spanning space exploration, first response, and defense, relies heavily on closing the loop between complex perception and reliable control. While advancements in deep learning have scaled the production and efficiency of autonomous vehicles, these systems still struggle with out-of-distribution scenarios, frequently requiring human-in-the-loop intervention to recover from navigation failures and stalls. To achieve persistent autonomy, the autonomy stack must generate the high-level reasoning typically provided by human supervisors.

The advent of Multimodal Large Language Models (MLLMs), such as Gemma 3 \cite{team2025kimi}, offers a mechanism to replace human oversight by leveraging web-scale, pre-trained knowledge for reasoning. Integrating MLLMs into the autonomy stack via Visual-Language-Action (VLA) models \cite{kawaharazuka2025vision} allows robots to process sensor observations alongside linguistic prompts to execute complex tasks. However, deploying VLAs currently presents two major bottlenecks: (1) a large computational overhead which clashes with the hardware constraints of mobile robots, and (2) the stochastic, ``black-box" nature of MLLMs, which compromises the deterministic safety required for real-time kinematic control.

Existing literature heavily relies on post-training, fine-tuning, or model distillation \cite{hinton2015distilling, kim2024openvla} to map large-scale public data to specific robot embodiments. Unfortunately, fine-tuning is computationally expensive, suffers from extreme data scarcity for niche applications, and struggles with ambiguous loss functions for generalized commands.

To address these limitations, we present WayFinder, an end-to-end hierarchical VLA framework that circumvents the fine-tuning bottleneck. The autonomy stack is separated into two distinct, asynchronous layers. The first is for high-level strategic reasoning, which is comprised of a pre-trained, zero-shot MLLM hosted offboard via edge computing, and is responsible for context-aware waypoint generation when the vehicle encounters complex obstacles. The second is for low-level, onboard kinematic control, which is comprised of a high-frequency policy that is dedicated to continuous collision avoidance and local point-to-point navigation.

By design, WayFinder relies on the safe and robust, low-level controller for the majority of operation. The computationally heavy MLLM is queried strictly as a recovery mechanism that is triggered only when the low-level policy detects when progress towards the target is stalled, such as the case when encountering a dead-end or infinite loop. This hierarchical framework leverages the advanced common-sense reasoning of MLLMs without the cost of fine-tuning, preserves the high-frequency safety guarantees of traditional controllers, and drastically minimizes inference costs. Through comprehensive simulation across varying map complexities, we demonstrate that WayFinder provides a scalable, reliable pathway to zero-shot robotic autonomy.

\subsection{Contributions}

\noindent
$\bullet$ We develop a Deep Reinforcement Learning (DRL) approach to train a high-frequency kinematic control policy. This policy efficiently navigates an autonomous robot between waypoints by fusing continuous feedback from onboard depth sensors, GPS, and IMU data.

\noindent
$\bullet$ We introduce a zero-shot integration of a high-level Multimodal Large Language Model (MLLM) for strategic waypoint generation and robust recovery from navigation failures. This includes an innovative method for formulating state maps as input to the MLLM, which better integrates with the VLM encoders utilized by most modern MLLMs, while exploring unknown space and building global maps.

\noindent
$\bullet$ We present WayFinder, an end-to-end autonomy stack that closes the loop between sensing and kinematic control for autonomous vehicles, by integrating the low-level and high-level policies into a hierarchical edge VLA framework that operates using the low-level policy by default and executes the high-level policy only when a ``stalled" state is detected and higher level reasoning is required to escape it.

\noindent
$\bullet$ We release the complete dataset used to train and evaluate the low-level DRL policy, alongside the full, open-source repository for executing the WayFinder hierarchical VLA pipeline, to support future research, at \footnote{https://github.com/WreckItTim/OmniNaviPy}.

We evaluate WayFinder in the high-fidelity Microsoft AirSim simulator across four maps of varying complexity. Although our framework is general and can be used in a broad range of settings, it is tested in AirSim thus we consider navigation from the perspective of airborne drones. To also assess the tradeoff between computational efficiency and navigational reliability, we evaluate the high-level policy using three different parameter scales of the Gemma 3 MLLM. Compared to baselines relying solely on low-level kinematic control, WayFinder significantly improves navigation success rates, varying by the environment and size of the MLLM  between 1.82\% to 27.45\%. This demonstrates an adaptable framework that can be tailored to specific edge-computing constraints without sacrificing recovery capabilities.

\section{RELATED WORK}


To overcome the limitations of rigid rule-based systems, Deep Reinforcement Learning (DRL) has emerged as a dominant paradigm for mapless navigation \cite{tai2017virtual}. By formulating navigation as a Markov Decision Process (MDP), DRL policies, such as Deep Q-Networks (DQN) \cite{mnih2015human}, can learn complex, non-linear mappings from high-dimensional sensor data (\textit{e.g.}, depth maps) directly to kinematic control commands \cite{johnsen2026cadence}; which also demonstrates the deployability onto real-world computing hardware. However, end-to-end DRL controllers face some mitigable shortcomings: they struggle when exposed to out-of-distribution scenarios (partially contributed to generalization error), and lack semantic ``common-sense" reasoning required to escape deadlocks, such as maze-like dead-ends or infinite loops.

The integration of MLLMs into robotics has led to the development of VLA models, such as RT-2 \cite{zitkovich2023rt} and OpenVLA \cite{kim2024openvla}. These frameworks process real-time visual observations alongside linguistic instructions to output robotic actions, offering unprecedented generalization for open-world, user-defined tasks. By leveraging the vast web-scale knowledge embedded in their pre-trained weights, VLAs exhibit improved precision in semantic reasoning and zero-shot task recognition compared to traditional DRL.

Despite these advantages, deploying VLAs in high-frequency, closed-loop mobile robotics presents severe challenges. The primary bottleneck is the large computational overhead required for continuous inference, which restricts their use on edge-computing hardware typical of lightweight robotics. Furthermore, translating high-level semantic reasoning into low-level continuous motor commands typically requires extensive fine-tuning (\textit{e.g.}, LoRA \cite{hu2022lora}). This fine-tuning exacerbates the issue of data scarcity, as large-scale, high-quality demonstration datasets are no only highly specific to particular robot embodiments, but are virtually non-existent for specialized aerial navigation tasks.

To circumvent the prohibitive costs of VLA fine-tuning, recent literature has explored zero-shot hierarchical frameworks. Works such as LM-Nav \cite{shah2023lm} utilize pre-trained language models as high-level semantic planners with fully observed global information that interface with pre-existing, low-level execution APIs. By decoupling the reasoning from the kinematics, these systems can generate waypoints without requiring embodiment-specific fine-tuning.

WayFinder directly advances this paradigm by applying it to high-frequency, closed-loop spatial navigation through global unknown environments and integrating recovery mechanisms to avoid human intervention when the robot becomes stalled. By translating local depth maps into a 3D deterministic, semantically grounded occupancy grid, WayFinder exploits the inherent tokenization strengths of Vision-Language Models (VLMs), while the robot is exploring an unknown environment and building a global representation of it. The generated global map further allows a zero-shot, offboard MLLM (Gemma 3) to act as an asynchronous recovery system that provides a solution to the deadlock shortcomings of DRL while completely bypassing the fine-tuning, data scarcity, and computational frequency constraints of traditional VLA architectures.

\section{SYSTEM FRAMEWORK}

Fig.~\ref{fig:wayfinder} illustrates the system framework of WayFinder. The low-level policy is a neural network which inputs locally observed 2D depth maps collected from onboard sensors into a Convolutional Neural Network (CNN)
, injects GPS and IMU data into the flattened layers of a Multi-Layer Perceptron (MLP)
, and outputs downstream motion actions for the drone to take. The low level policy network is trained using DRL in the robust drone simulator Microsoft AirSim. 

After DRL training and during deployment, we integrate a component to supervise progression of the drone towards its given target location. A simple boolean is measured based on the Euclidean progress towards the target within a given viewing window, and if not enough progress is made then a stalled state is triggered. Also during deployment, we consider that the onboard sensor data is being relayed to a compute-capable device (such as an edge server). The relayed depth, GPS, and IMU data is leveraged to build a world map with visual representations of the drone, its current and past trajectory, the start and target positions, and detected obstacles in the form of an occupancy grid.

Upon activation of the trigger, the high-level policy is queried in an attempt to make more significant progress.  The current world map is paired with linguistic commands, along with added context, providing instructions to generate a new waypoint which will help recover the drone from its stalled state. The visual-language data is fed into an MLLM, of which the waypoint coordinates are extracted and set as the new target for the drone to navigate to. After reaching a waypoint, the target is set to the initial goal and the drone continues its path unless its progress is stalled again.

\section{LOW-LEVEL POLICY VIA DRL}

The fundamental navigation problem addressed by the low-level policy is localized, point-to-point traversal through unknown environments to a given target coordinate, which can be either a global mission objective or an intermediate waypoint generated by the high-level policy. The autonomous agent must execute kinematic maneuvers to avoid uncharted obstacles. This micro-navigation requires a responsive, computationally lightweight controller capable of processing high-dimensional egocentric sensor data directly onboard embedded hardware. Within the WayFinder framework, this reactive policy serves as the default operational mode, managing all routine flight dynamics and immediate collision avoidance so that the computationally expensive MLLM is reserved exclusively for computationally heavy reasoning.

Baseline operations of the WayFinder framework are executed by this high-frequency, reactive kinematic controller and is trained via DRL. Since the kinematic action space $\mathcal{A}$ is discrete (parameterized into specific translational movements and yaw rotations), we formulate the policy using a Deep Q-Network (DQN) \cite{mnih2015human} architecture. The local navigation problem is framed as a Partially Observable Markov Decision Process (POMDP), defined by the tuple $(\mathcal{S}, \mathcal{A}, \mathcal{T}, \mathcal{R})$.

\subsection{State Space $\mathcal{S}$}

To capture dynamic environmental changes, the state space $\mathcal{S}$ utilizes a sliding temporal window of size $\tau$. As depicted in the system architecture (Fig.~\ref{fig:wayfinder}), the state fuses high-dimensional exteroceptive depth sensors with low-dimensional proprioceptive data:
\begin{equation}
\label{eq:st}
\begin{aligned}
S_t = \Big\{ \mathbf{D}_{k}, \mathbf{c}_{k}, \theta_{k} \Big\}_{k=t-\tau}^{t}
\end{aligned}
\end{equation}
where $\mathbf{D} \in \mathbb{R}^{144 \times 256}$ denotes the 2D depth map, $\mathbf{c} \in \mathbb{R}^2$ denotes the local GPS coordinates relative to either the target or current waypoint, and $\theta \in \mathbb{R}^1$ represents the IMU heading. Note that we use relative GPS coordinates to not overfit to specific coordinates on the map during DRL training, which would otherwise be the case for absolute GPS coordinates.

\subsection{Action Space $\mathcal{A}$}

The action space $\mathcal{A}$ is defined as a discrete set of kinematic commands. Specifically, the policy selects from a finite set of translational movements (meters) and yaw rotations (degrees) relative to the robot's current pose. This discretization drastically reduces the dimensionality of the space needed to explore for DRL, facilitating highly sample-efficient convergence while maintaining precise control. Execution of these discrete commands into continuous translational and rotational accelerations, as required for stable flight, is accomplished by an assumed low-level onboard hardware flight controller -- which can vary by the given robot's firmware.

\subsection{Transition Function $\mathcal{T}$ and Collision Avoidance} 

The transition function $\mathcal{T}(s_{t+1}|s_t, a_t)$ dictates how the state space changes given an action. The base kinematics, physics, and graphics are rendered by the Microsoft AirSim simulator and determine how the environment transitions from one state to the next, by simulating the low-level onboard hardware flight controller to move a drone by the given rotational or translational action $a_t$. We introduce a deterministic collision avoidance algorithm that considers the vehicle is equipped with a short-range distance sensor. If an obstacle is detected during execution of the action, then the physical movement is halted before a collision can occur, and a new state $s_{t+1}$ is sampled from the halted position.

\subsection{Reward Function $\mathcal{R}$} 
Since physical collisions are prevented by the collision avoidance mechanism, they are entirely removed from the POMDP's penalty space. The reward function $r_t$ is therefore streamlined to focus strictly on path efficiency and temporal constraints. The episode terminates either when the goal is reached or a maximum step limit, $L$, is exceeded:
\begin{equation}
\label{eq:rt}
\begin{aligned}
r_t = \lambda_p d_{t} + \lambda_t r_{step} + r_{terminal}
\end{aligned}
\end{equation}
Where $d_{t}$ is used to encourage movement toward the immediate target, defined by the Euclidean distance: 
\begin{equation}
\label{eq:rprogress}
\begin{aligned}
d_{t} = \|\mathbf{p}_{t} - \mathbf{g}_{t}\|_2 
\end{aligned}
\end{equation}
Where $\mathbf{p}_t \in \mathbb{R}^2$ denotes the absolute position of the autonomous robot at time step $t$, and $\mathbf{g}_t \in \mathbb{R}^2$ denotes the target's absolute position. The time penalty ($r_{step}$) is a constant negative value applied at every timestep. If the robot selects an action that triggers the collision avoidance mechanism, it is halted and forced to take another timestep, inherently penalizing the inefficient, blocked action via this step penalty. A terminal reward/penalty ($r_{terminal}$) is applied at the end of each episode, and yields a large positive scalar if the vehicle reaches the target waypoint radius, or a heavy negative penalty if the episode is terminated due to exceeding a maximum number of steps $L$, discouraging infinite loops.

\subsection{Network Architecture}
\begin{figure*}[thpb]
  \centering
  \includegraphics[scale=0.8]{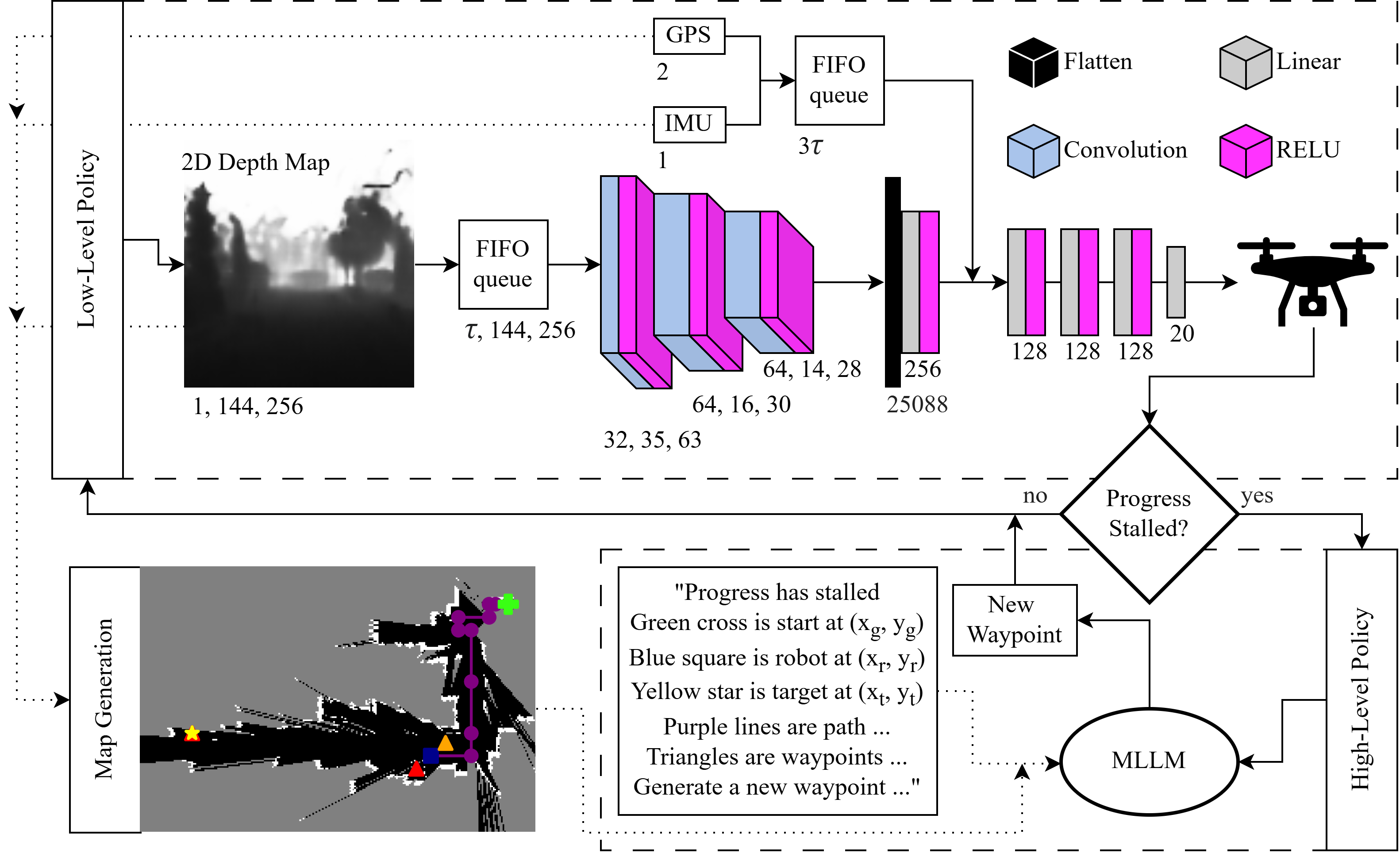}
  \caption{WayFinder: closed-loop autonomy that adjusts kinematic control in response to sensors, querying an MLLM for waypoint generation when stalled.}
  \label{fig:wayfinder}
\end{figure*}

The network architecture is illustrated along with the overall system model in Fig.~\ref{fig:wayfinder}. The DQN processes the multimodal state $S_t$ through a two-stage injection forward pass. A temporal queue of depth maps is fed through a CNN backbone consisting of three convolutional layers (with ReLU activations) to extract spatial features. The output is flattened into a 25,088-dimensional vector and concatenated with the flattened $\tau=3$ FIFO queue containing the GPS and IMU history. This fused representation is then passed through a MLP consisting of three 128-node linear layers, outputting the estimated Q-values, $Q(S_t, a_t)$, for each discrete kinematic action. The general structure is inspired by architectures used for learning Atari environments \cite{mnih2015human}, though the specific hyper parameters were explored with a grid search to find optimal settings.

\subsection{State Caching and Discretization} 

Training vision-based DRL policies in high-fidelity simulators is traditionally bottlenecked by rendering latencies. To ensure robust convergence while maintaining computational feasibility, we bypass the computational overhead of synchronous rendering by precomputing and caching the environment interactions. Prior to training, we discretize the navigable space of the AirSim environments, collecting and storing all possible egocentric depth maps alongside their corresponding Boolean collision states. Sampling directly from this discretized state space reduces the total training time of the DQN from several weeks to mere hours. 

\subsection{Offline A* Path Generation} 

During the pre-computing phase, we generate a comprehensive dataset of ground-truth navigation routes using an offline A* search algorithm \cite{hart1968formal}. The A* algorithm is utilized strictly to quantify the complexity of each path, defining the ``difficulty" of a path by the total number of optimal steps required to reach the target from the spawn point. This offline dataset is subsequently partitioned into a large training set and two static hold-out sets reserved for validation and testing (which are generated from regions of the AirSim environment unseen during training).

\subsection{Curriculum Learning and Replay Buffer} 

During the DRL training phase, the robot is procedurally exposed to paths of varying complexity via a curriculum learning schedule \cite{bengio2009curriculum}. The policy first learns simple paths with low A* step counts before progressively advancing to more difficult paths with higher step counts. However, sequential curriculum learning often leaves neural networks vulnerable to catastrophic forgetting. To mitigate this, we utilize a high-capacity replay buffer (allocating 100 GB of RAM). We also employ a stochastic sampling strategy, so that while the robot primarily trains on the current difficulty, paths from previously learned difficulties are also sampled from at a lower rate. Finally, we employ an early stopping mechanism that evaluates the policy against the static A* validation set, freezing the weights that yield the highest validation accuracy to serve as the final low-level model.

\section{HIGH-LEVEL POLICY VIA MLLM}

When the trigger condition $T_t = \text{True}$ is satisfied, the system halts the low-level kinematic policy and constructs a multimodal query to interface with the offboard MLLM. This query consists of two primary components: a synthesized visual state map with clearly indicated key features, and a structured linguistic prompt with commands and context that links visual features to precise coordinates.

An intuitive alternative is to establish a deterministic mapping between specific spatial coordinates and pixel indices within the image array. However, state-of-the-art MLLMs such as Gemma 3 frequently employ dynamic rescaling and tiling techniques during visual pre-processing, such as the Pan \& Scan (P\&S) algorithm as inspired by LLaVA \cite{liu2023visual}, to handle varying input resolutions. This dynamic rescaling breaks the mapping between pixel indices and spatial coordinates, leading to severe precision loss when generating waypoints. This mapping is further complicated by VLM encoders, such as SigLIP \cite{zhai2023sigmoid}, which tokenize the input.

\subsection{Triggering a Query} 

Querying the MLLM at every time step is both ineffective (causing hallucinations that deviate away from good trajectories) and inefficient (causing a high computational overhead). Thus we implement a sparse trigger condition that evaluates when the progress towards the target is stalled, such as when the robot is stuck in a dead end or an infinite loop. The vehicle's maximum forward progress toward the target is evaluated over a rolling discrete window of $N$ waypoints. At the current time step $t$, the path history is analyzed over a temporal window spanning from step $t-N$ to $t$. The spatial progress metric, $\Delta \rho_t$, is calculated as the difference between the initial distance to the goal at the start of the window ($d_{t-N}$) and the absolute minimum distance to the goal achieved at any point within that window:
\begin{equation}
\label{eq:deltarhot}
\begin{aligned}
\Delta \rho_t = d_{t-N} - \min_{k \in [t-N, t]} (d_k).
\end{aligned}
\end{equation}
The system requires a minimum history of $N$ waypoints before evaluation can begin. Once populated, the high-level MLLM policy is invoked if the maximum forward progress falls below a predefined stagnation threshold, $\epsilon$. The trigger condition $T_t$ is formally defined as a Boolean status:
\begin{equation}
\label{eq:tt}
\begin{aligned}
T_t = \begin{cases} \text{True}, & \text{if } (\Delta \rho_t < \epsilon) \\ \text{False}, & \text{otherwise}. \end{cases}
\end{aligned}
\end{equation}
When $T_t = \text{True}$, the system flags a stalled status, the low-level policy is temporarily suspended, the robot's current state maps and linguistic context are transmitted to the offboard MLLM to generate a restorative waypoint, and the temporal window is reset to not trigger a subsequent false positive when detecting a stalled status.

\subsection{Visual State Map}

Instead of passing raw, high-dimensional sensor data (\textit{i.e.}, RGB-D) directly to the MLLM, WayFinder abstracts the perceived environment into a  state map which is a 2D aerial view represented in RGB color space, $M_t \in \mathbb{R}^{H \times W \times 3}$. 

This design choice is specifically tailored to the architecture of contemporary MLLMs, such as Gemma 3 \cite{team2025kimi}, which utilize VLMs like SigLIP \cite{zhai2023sigmoid} to extract visual tokens. By projecting obstacles and mission-critical coordinates using highly distinct RGB color channels and primitive geometric shapes, we optimize the visual feature extraction process. These stark contrast features are more efficiently tokenized by the VLM's visual encoder, ensuring that high-fidelity spatial representations are passed to the language model. The semantic markers projected onto $M_t$ include, in bottom-to-top order of projection onto the map:

\begin{enumerate}
    \item Free space is indicated by black pixels, which represents traversable space in a binarized occupancy grid.
    \item Obstacles are indicated by white pixels, which represents non-traversable space.
    \item Undetermined space is indicated by grey pixels, which represents unexplored or uncertain regions which may or may not be traversable.
    \item The robot's historical trajectory taken by the robot up to time $t$, is indicated by purple lines and circles.
    \item Any previously generated waypoints are indicated by red triangles, which provides context into attempted restorations in the past, for higher level reasoning.
    \item The start position at $t=0$ is shown by a green cross. 
    \item The robot's position $\mathbf{p}_t$ is indicated by a blue square.
    \item The global target $\mathbf{g}$ is indicated by a yellow star.
\end{enumerate}

\subsection{Linguistic Text Prompt}

To bridge the visual representation with semantic reasoning, $M_t$ is paired with a structured text prompt. Since MLLMs rely on cross-modal attention mechanisms to fuse visual tokens with text, pairing the distinct shapes and colors in $M_t$ with explicit linguistic definitions and spatial coordinates forces a strong semantic alignment. A standard query follows the template in the following paragraph.

``You are the supervisor for a robot tasked to navigate to a target position. The robot can only see what is in front of it, but you have access to the full map and path history. The robot can navigate on its own, but it has just gotten stuck on a complex set of obstacles and is asking for your help to generate an intermediate waypoint to get unstuck and make progress towards the target. The robot can only make discrete movements forward of between 1 to 32 meters or rotate the direction it is facing by 90 degrees. See the given image for reference, which is an illustration of the current map and robot state. The provided image was created by updating an occupancy grid through ray tracing of the robots egocentric depth sensors, IMU, and GPS, thus is only partially observed. The robot (blue square) is at ($x_r$, $y_r$, $\theta_r$) as measured in (meters, meters, degrees). The target (yellow star) is at ($x_t$, $y_t$). The start (green cross) is at ($x_o$, $y_o$). The robot has traveled with a path history (purple lines and circles) of (x, y, theta) coordinates: [($x_1$, $y_1$, $\theta_1$), ... , ($x_n$, $y_n$, $\theta_n$)]. Consider the previous attempted waypoints (red triangles), which were generated to try and unstuck the robot to no avail, are at the following (x, y) coordinates: [($x_1$, $y_1$), ... , ($x_m$, $y_m$)]. White pixels are obstacles. Black pixels are safe. Gray pixels are unknown. Only generate a waypoint at a safe (black space) (x, y) coordinate, and one that has not been attempted yet. Identify any complex obstacles that may block the robot from reaching the target. Specifically format your response as: [STRATEGY]: reason for generating the waypoint. [WAYPOINT]: (x, y). Do not include any conversational filler."

By explicitly defining the geometric relationships in text, we ground the MLLM's visual encoder, reducing spatial ambiguity. The pre-trained MLLM (e.g., Gemma 3) processes this multimodal input pair and generates a new spatial coordinate. This output acts as a strategic intermediate waypoint, routing the robot out of the current deadlock scenario. Once generated, this new waypoint replaces the immediate goal for the low-level policy, the trigger condition $T_t$ is reset to False, and high-frequency onboard kinematic control resumes. 

\subsection{Ray Tracing to Build Global Maps}
\label{sec:ray}

To construct the global map ($M_t$) required by the MLLM, the robot's egocentric depth observations must be projected into a global spatial representation. We achieve this by transforming raw 2D depth maps into a discrete ternary occupancy grid, where each cell represents one of three states: free ($0$), occupied ($1$), or unknown ($-1$).

Given a depth map $\mathbf{D} \in \mathbb{R}^{144 \times 256}$, we first map each pixel $(u, v)$ to its corresponding 3D coordinate $(x, y, z)$ in the camera frame. Using the assumed camera intrinsic parameters, focal lengths $(f_x, f_y)$ and principal points $(c_x, c_y)$, we compute the normalized coordinates:
\begin{equation}
\label{eq:xnorm}
\begin{aligned}
x_{norm} = \frac{u - c_x}{f_x}, \quad y_{norm} = \frac{v - c_y}{f_y}
\end{aligned}
\end{equation}
To account for the spherical distortion of the depth sensor, we calculate the exact ray length for each pixel and project the depth along the optical axis ($z$):
\begin{equation}
\label{eq:z}
\begin{aligned}
z = \frac{\mathbf{D}(u, v)}{\sqrt{x_{norm}^2 + y_{norm}^2 + 1}}
\end{aligned}
\end{equation}
The full 3D Cartesian coordinates are then recovered via $x = x_{norm} \cdot z$ and $y = y_{norm} \cdot z$.

We apply a spatial filter along the vertical $Y$-axis, where points outside of a given range $[y_{min}, y_{max}]$ are discarded, effectively ignoring the ground and high-hanging obstacles (\textit{e.g.}, telephone wires) that do not impede the vehicle's current trajectory. We also apply a horizon threshold on the depth ($z$) to not see outside the predefined bounds of the map. The filtered 3D points are projected onto a local $XZ$-plane.

A local occupancy grid is created by comparing the depth of each cell ($z_{grid}$) against the detected obstacle depth ($z_{obs}$) of the ray that passes through the cell. A cell is marked as free if it is in front of the nearest obstacle ($z_{grid} < z_{obs} - \delta_{occ}$), as occupied if its depth falls within a tolerance of the obstacle ($|z_{grid} - z_{obs}| \le \delta_{occ}$), and those lying behind an obstacle ($z_{grid} > z_{obs} + \delta_{occ}$) remain initialized as unknown.

The resulting occupancy grid is then rotated by the robot's heading $\theta_t$ and translated by the robot's position $\mathbf{p}_t$, and then used to update the world map $M_t$. A masking operation ensures that only newly observed free or occupied cells overwrite the world map, preserving the memory of previously explored regions while updating it with more recent ones.

\section{RESULTS}

\subsection{Experimental Setup}

To evaluate the recovery capabilities and computational efficiency of the WayFinder framework, we conduct closed-loop navigation experiments within Microsoft AirSim, a high-fidelity physics and graphics simulator. Our evaluation isolates the impact of the asynchronous high-level policy against only using the low-level policy, with design considerations in the parameter space.

We evaluate the framework across four distinct AirSim environments, as released by Microsoft, selected to represent a progressive scale of complexity. The ``Blocks" environment is of the lowest complexity, featuring sparse simple geometric obstacles and wide line-of-sight. ``AirSimNH" is an urban environment with shallower line of sight, and with more dense and complex objects such as houses, roads, trees, and cars. ``CityEnviron" is a dense metropolitan map also featuring complex intersections and larger buildings. ``Coastline" is the largest map with open oceans, and a road cutting through the coast with trees, rocks, and small islands.

To ensure unbiased evaluation, we utilize the offline A* algorithm to generate a static, hold-out testing set of 4,000 paths. The A* algorithm verifies that a valid route exists for every path in the test set. Performance is quantified by the navigation accuracy, which is the percentage of test episodes where the robot successfully navigates to within the target radius before exceeding the maximum step limit, $L$. We further mitigate statistical variance by using a constant random seed during evaluation, and setting the ``temperature" parameter of the MLLM to zero.

One objective of WayFinder is to mitigate the computational bottleneck typically associated with VLA models, to deploy to edge devices. To quantify the tradeoff between zero-shot spatial reasoning and inference latency, we benchmark the high-level policy using three distinct parameter scales of the Gemma 3 architecture (4B, 12B, and 27B).

\subsection{Fully Observed Map}

We first investigate navigation accuracy as a function of the Gemma 3 model size and the AirSim environment. In this baseline experiment, the entire map is fully observable to the high-level policy. We utilize the ground truth occupancy grid directly from AirSim, bypassing the ray-tracing component, to establish an upper performance bound for the MLLM. We also establish a baseline performance by using only the low-level policy for navigation. Table~\ref{tab:full_results} lists the results.

\begin{table}[t]
\caption{Navigation accuracies using the fully observed map.}
\begin{center}
\begin{tabular}{|c||c|c|c|c|}
\hline
MLLM model & AirSimNH & Blocks & Coastline & CityEnviron \\ 
\hline
none & 84.91\% & 83.71\% & 63.73\% & 44.55\% \\
gemma3:4b & 92.91\% & 89.57\% & 69.64\% & 57.91\% \\
gemma3:12b & 94.55\% & 88.71\% & 73.00\% & 64.27\% \\
gemma3:27b & 96.09\% & 92.43\% & 77.73\% & 70.73\% \\
\hline
\end{tabular}
\label{tab:full_results}
\end{center}
\end{table}

The low-level DQN policy was trained exclusively within the AirSimNH environment. Consequently, the baseline accuracy, labeled as ``none" in table~\ref{tab:full_results}, drops significantly in the other three environments due to out-of-distribution generalization errors. The integration of the high-level MLLM policy successfully mitigates this overfitting, and overall yields substantial accuracy improvements, establishing its efficacy in resolving deployment errors. The accuracy increases monotonically with model size across all environments, highlighting a clear tradeoff between computational efficiency and navigational reliability. 

\subsection{Recovery Rates}

While the asynchronous stall detection mechanism is robust, it remains susceptible to false positives, occasionally triggering the MLLM when the robot is navigating around large obstacles. This can be mitigated by expanding the temporal window ($N$) used to calculate progress; however, doing so consequently delays stall detection and requires relaxation of the maximum time limit, $L$. Since the optimal value of $N$ is a function of specific mission requirements, we isolate the performance of the MLLM by directly measuring its recovery rate. This metric is defined as the percentage of episodes that successfully reached the target after initially failing under the sole execution of the low-level policy. Conversely, failed recovery rates track instances where MLLM-generated waypoints actively diverted the robot away from a viable trajectory or introduced new failure modes. 

Table~\ref{tab:full_results_recovery} and Table~\ref{tab:full_results_failed} present the successful and failed recovery rates. Scaling the MLLM parameter size not only increases the successful recovery rate but also decreases the failed recovery rate. Notably, failure rates are most pronounced in the Coastline environment, which inherently has the largest distribution shift from the AirSimNH environment, the largest spatial area, and most open space.

\begin{table}[t]
\caption{Recovery rates using the fully observed map.}
\begin{center}
\begin{tabular}{|c||c|c|c|c|}
\hline
MLLM model & AirSimNH & Blocks & Coastline & CityEnviron \\ 
\hline
gemma3:4b & +59.64\% & +35.96\% & +33.08\% & +25.57\% \\
gemma3:12b & +67.47\% & +31.58\% & +40.85\% & +37.70\% \\
gemma3:27b & +77.71\% & +53.51\% & +52.38\% & +48.03\% \\
\hline
\end{tabular}
\label{tab:full_results_recovery}
\end{center}
\end{table}

\begin{table}[t]
\caption{Failed recovery rates using the fully observed map.}
\begin{center}
\begin{tabular}{|c||c|c|c|c|}
\hline
MLLM model & AirSimNH & Blocks & Coastline & CityEnviron \\ 
\hline
gemma3:4b & -1.18\% & -0.00\% & -9.56\% & -1.84\% \\
gemma3:12b & -0.64\% & -0.17\% & -8.70\% & -2.65\% \\
gemma3:27b & -0.64\% & -0.00\% & -7.85\% & -1.02\% \\
\hline
\end{tabular}
\label{tab:full_results_failed}
\end{center}
\end{table}

\subsection{WayFinder -- Partially Observed Map}

Having established the theoretical upper bound using fully observed maps, we next evaluate WayFinder under practical edge-deployment constraints. In this configuration, the MLLM relies entirely on the global map $M_t$ dynamically constructed via the agent's depth sensor and ray-tracing pipeline detailed in Sec.~\ref{sec:ray}. This results in a partially observed, uncertain representation of the environment.

We investigate using the global maps generated via ray tracing, see Sec.~\ref{sec:ray}, which result in only a partial and uncertain representation of the environment to be used by the high-level policy. As opposed to using the ground truth occupancy grid as in the results presented in table~\ref{tab:full_results_recovery}, those in table~\ref{tab:partial_results_recovery} list the recovery rates for using the partially observed global map $M_t$ with the ray-traced occupancy grid. Note that we compare recovery rates going forward for two reasons: (1) the failed recovery rates can be mitigated by improving trigger conditions to query the MLLM, and (2) evaluating against the full sample size (4400 paths) takes almost a full day to compute, however only computing the recovery rate significantly reduces the sample size (to 1289 paths) and brings compute time down to hours. The overall accuracies range between 44.55-83.71\% when not using the MLLM and 72.00-96.64\% with it.

\begin{table}[t]
\caption{Recovery rates using the partially observed map.}
\begin{center}
\begin{tabular}{|c||c|c|c|c|}
\hline
MLLM model & AirSimNH & Blocks & Coastline & CityEnviron \\ 
\hline
gemma3:4b & +54.82\% & +28.95\% & +25.81\% & +25.08\% \\
gemma3:12b & +65.06\% & +36.84\% & +41.60\% & +36.07\% \\
gemma3:27b & +80.72\% & +64.91\% & +53.63\% & +50.82\% \\

\hline
\end{tabular}
\label{tab:partial_results_recovery}
\end{center}
\end{table}

Table~\ref{tab:partial_results_recovery} shows that using the partially observed map results in small changes in recovery rates, with the smallest drops in the recovery rates coming from the largest Gemma 3 model (27b). This demonstrates the robustness of the partial maps, and that the full map outside of view of the robot is not as advantageous to generating recovery waypoints. The recovery rates slightly improve for some maps, and we go into more detail about this phenomenon in the next section.

\subsection{Ablation Study}

We conduct an ablation study to evaluate the contribution of individual components within the WayFinder framework, specifically focusing on the multimodal prompt formulation. Table \ref{tab:ablation} reports the aggregate recovery rates and relative path lengths across all AirSim environments for various system configurations. Unless otherwise specified, the default configuration utilizes the Gemma 3 (27B) model, partially observed egocentric maps ($M_t$), and comprehensive visual-language inputs (including path and waypoint histories). The relative path length is defined as the average ratio, calculated over all successful episodes, of the total steps required to reach the target under a specific configuration compared to the default WayFinder setup. This metric is critical because it directly correlates with navigational efficiency, which is paramount for mission success in edge robotics where strict time, power, and energy constraints apply. 

Configuation \textit{(I)} establishes the baseline performance of the default WayFinder framework. We then systematically modify or ablate components of the prompt to isolate their impact. These independent modifications include: \textit{(II-III)} scaling down the MLLM parameter size; \textit{(IV)} omitting explicit spatial coordinates from the linguistic prompt, forcing the model to rely entirely on the visual state map guided only by drawn $x$ and $y$ axes with labeled tick marks; \textit{(V)} removing the waypoint history, preventing the model from recalling previously failed recovery attempts; \textit{(VI)} removing the visual state map $M_t$, forcing the model to rely strictly on linguistic, coordinate-based instructions; \textit{(VII)} removing the historical trajectory, blinding the MLLM to previously traveled paths; and \textit{(VIII)} replacing the ray-traced maps with ground-truth occupancy grids extracted directly from AirSim's voxel data.

\begin{table}[t]
\caption{Ablation study, showing the total recovery rate and relative path length over all AirSim environments.}
\begin{center}
\begin{tabular}{|c|c|c|}
\hline
Configuration & Recovery Rate & Rel. Length \\ 
\hline
\textit{(I)} WayFinder (default settings) & 56.79\% & 1.00 \\
\textit{(II)} Using Gemma3:4b & 30.88\% & 1.17 \\
\textit{(III)} Using Gemma3:12b & 41.74\% & 1.17 \\
\textit{(IV)} No Coordinates in Language & 12.26\% & 1.46 \\
\textit{(V)} No Waypoint History & 48.02\% & 0.99 \\
\textit{(VI)} No Visual Component & 55.62\% & 1.10 \\
\textit{(VII)} No Path History & 52.37\% & 1.12 \\
\textit{(VIII)} Ground Truth Occupancy Grid & 53.45\% & 1.05 \\
\hline
\end{tabular}
\label{tab:ablation}
\end{center}
\end{table}

Configurations \textit{II-III} warrant that performance is strongly correlated with the MLLM parameter size, demonstrating that advanced reasoning directly scales with model size.

There is a large degradation in performance resulting from omitting the explicit linguistic coordinates, as in configuration \textit{IV}, forcing the MLLM to rely solely on the coordinate system drawn onto $M_t$. This generates more erratic waypoints that are further from the agent's position, and warrants that linking linguistic coordinates to visual map features is a robust prompt engineering technique. We posit that this is likely a general trend, that can be applied to many types of spatial reasoning problems.

We find that omitting the waypoint history can trap the agent in an infinite loop, as the ungrounded MLLM repeatedly generates the same failed waypoint without memory of prior attempts. Configuration \textit{V} reflects this with a corresponding drop in the evaluated recovery rate.

Interestingly, removing the global map $M_t$, as done in configuration \textit{VI}, only moderately degrades performance. We posit that the MLLM is capable of extrapolating global spatial structures from the linguistic coordinate arrays provided for the start, current, target, and historical path positions. This is further warranted by the degradation in performance that results from removing the historical path points from the language prompt (configuration \textit{VII}).

Counterintuitively, substituting the ray-traced maps with perfect, ground-truth occupancy grids from AirSim, configuration \textit{VIII}, results in a performance degradation. The ray-traced maps provide a localized, partially observed view that reflects the explored vicinity. We posit that providing the localized maps, as opposed to ground truth global ones, explicitly forces the MLLM to reason based on the robot's perspective thus allowing for egocentric-based logic. However, only the largest model resulted in consistent improvements.



\section{CONCLUSIONS}

In this paper, we presented WayFinder, an end-to-end hierarchical Vision-Language-Action (VLA) framework designed for resilient navigation in unknown environments. At the foundation of the system is a lightweight, high-frequency Deep Reinforcement Learning (DRL) policy that executes localized kinematic control and deterministic collision avoidance, achieving baseline navigation accuracies of between 44.55-83.71\% (considering out-of-distribution and generalization errors when evaluating on novel and complex environments, especially those not seen during training). To overcome the susceptibility of purely reactive controllers to complex and out-of-distribution scenarios, a Multimodal Large Language Model (MLLM) is asynchronously queried upon detecting stalled progress. By processing a global map with structured linguistic context, the high-level policy generates recovery waypoints, effectively routing the robot out of stalled scenarios, improving the overall navigation accuracy to 72.00-96.64\%. Thus, WayFinder is an effective means for overcoming shortcomings of DRL-based controllers by providing a low-frequency supervisor to help navigate away from out-of-distribution dead ends (including those arising from generalization error) and around complex obstacles. 

We posit that these methods can be applied to a variety of low-level controllers, other than just DRL-based ones, primarily because WayFinder operates the MLLM entirely zero-shot. By querying it strictly as a failure-recovery mechanism, WayFinder successfully circumvents the prohibitive data-scarcity and computational fine-tuning bottlenecks typical of conventional VLA architectures. The hierarchical decoupling demonstrated in this framework can thus be generalized to rapidly adapt a wide variety of existing autonomous robotic systems, significantly enhancing their long-horizon reliability and spatial reasoning capabilities without the need for embodiment and environment-specific learning. 

\addtolength{\textheight}{-12cm}   




\section*{ACKNOWLEDGMENTS}

This work was partially supported by the U. S. National
Science Foundation under Grant No. CCF-2140154.


\bibliographystyle{IEEEtran}
\bibliography{bibliography}

@IEEEtranBSTCTL{IEEEexample:BSTcontrol,
  CTLuse_forced_etal       = "yes",
  CTLmax_names_forced_etal = "3",
  CTLnames_show_etal       = "1"
}

@article{johnsen2026cadence,
  title={CADENCE: Context-Adaptive Depth Estimation for Navigation and Computational Efficiency},
  author={Johnsen, Timothy K and Levorato, Marco},
  journal={arXiv preprint arXiv:2604.07286},
  year={2026}
}

@article{team2025kimi,
  title={Kimi-vl technical report},
  author={Team, Kimi and Du, Angang and Yin, Bohong and Xing, Bowei and Qu, Bowen and Wang, Bowen and Chen, Cheng and Zhang, Chenlin and Du, Chenzhuang and Wei, Chu and others},
  journal={arXiv preprint arXiv:2504.07491},
  year={2025}
}

@inproceedings{zhai2023sigmoid,
  title={Sigmoid loss for language image pre-training},
  author={Zhai, Xiaohua and Mustafa, Basil and Kolesnikov, Alexander and Beyer, Lucas},
  booktitle={Proceedings of the IEEE/CVF international conference on computer vision},
  pages={11975--11986},
  year={2023}
}

@article{liu2023visual,
  title={Visual instruction tuning},
  author={Liu, Haotian and Li, Chunyuan and Wu, Qingyang and Lee, Yong Jae},
  journal={Advances in neural information processing systems},
  volume={36},
  pages={34892--34916},
  year={2023}
}

@article{hinton2015distilling,
  title={Distilling the knowledge in a neural network},
  author={Hinton, Geoffrey and Vinyals, Oriol and Dean, Jeff},
  journal={arXiv preprint arXiv:1503.02531},
  year={2015}
}

@article{kawaharazuka2025vision,
  title={Vision-language-action models for robotics: A review towards real-world applications},
  author={Kawaharazuka, Kento and Oh, Jihoon and Yamada, Jun and Posner, Ingmar and Zhu, Yuke},
  journal={IEEE Access},
  year={2025},
  publisher={IEEE}
}

@article{kim2024openvla,
  title={Openvla: An open-source vision-language-action model},
  author={Kim, Moo Jin and Pertsch, Karl and Karamcheti, Siddharth and Xiao, Ted and Balakrishna, Ashwin and Nair, Suraj and Rafailov, Rafael and Foster, Ethan and Lam, Grace and Sanketi, Pannag and others},
  journal={arXiv preprint arXiv:2406.09246},
  year={2024}
}

@inproceedings{tai2017virtual,
  title={Virtual-to-real deep reinforcement learning: Continuous control of mobile robots for mapless navigation},
  author={Tai, Lei and Paolo, Giuseppe and Liu, Ming},
  booktitle={2017 IEEE/RSJ international conference on intelligent robots and systems (IROS)},
  pages={31--36},
  year={2017},
  organization={IEEE}
}

@article{mnih2015human,
  title={Human-level control through deep reinforcement learning},
  author={Mnih, Volodymyr and Kavukcuoglu, Koray and Silver, David and Rusu, Andrei A and Veness, Joel and Bellemare, Marc G and Graves, Alex and Riedmiller, Martin and Fidjeland, Andreas K and Ostrovski, Georg and others},
  journal={nature},
  volume={518},
  number={7540},
  pages={529--533},
  year={2015},
  publisher={Nature Publishing Group}
}

@inproceedings{zitkovich2023rt,
  title={Rt-2: Vision-language-action models transfer web knowledge to robotic control},
  author={Zitkovich, Brianna and Yu, Tianhe and Xu, Sichun and Xu, Peng and Xiao, Ted and Xia, Fei and Wu, Jialin and Wohlhart, Paul and Welker, Stefan and Wahid, Ayzaan and others},
  booktitle={Conference on Robot Learning},
  pages={2165--2183},
  year={2023},
  organization={PMLR}
}

@article{hu2022lora,
  title={Lora: Low-rank adaptation of large language models.},
  author={Hu, Edward J and Shen, Yelong and Wallis, Phillip and Allen-Zhu, Zeyuan and Li, Yuanzhi and Wang, Shean and Wang, Liang and Chen, Weizhu and others},
  journal={Iclr},
  volume={1},
  number={2},
  pages={3},
  year={2022}
}

@inproceedings{shah2023lm,
  title={Lm-nav: Robotic navigation with large pre-trained models of language, vision, and action},
  author={Shah, Dhruv and Osi{\'n}ski, B{\l}a{\.z}ej and Levine, Sergey and others},
  booktitle={Conference on robot learning},
  pages={492--504},
  year={2023},
  organization={pmlr}
}

@article{hart1968formal,
  title={A formal basis for the heuristic determination of minimum cost paths},
  author={Hart, Peter E and Nilsson, Nils J and Raphael, Bertram},
  journal={IEEE transactions on Systems Science and Cybernetics},
  volume={4},
  number={2},
  pages={100--107},
  year={1968},
  publisher={IEEE}
}

@inproceedings{bengio2009curriculum,
  title={Curriculum learning},
  author={Bengio, Yoshua and Louradour, J{\'e}r{\^o}me and Collobert, Ronan and Weston, Jason},
  booktitle={Proceedings of the 26th annual international conference on machine learning},
  pages={41--48},
  year={2009}
}

\end{document}